# Coregionalised Locomotion Envelopes – A Qualitative Approach

**Neil Dhir**[*]
Department of Engineering Science
University of Oxford
`neild@robots.ox.ac.uk`

**Houman Dallali**
Department of Computer Science
California State University
Channel Islands, CA, USA

**Mo Rastgaar**
Department of Mechanical Engineering
Michigan Technological University
MI, USA

## Abstract

'Sharing of statistical strength' is a phrase often employed in machine learning and signal processing. In sensor networks, for example, missing signals from certain sensors may be predicted by exploiting their correlation with observed signals acquired from other sensors [1]. For humans, our hands move synchronously with our legs, and we can exploit these implicit correlations for predicting new poses and for generating new natural-looking walking sequences. We can also go much further and exploit this form of transfer learning, to develop new control schemas for robust control of rehabilitation robots. In this short paper we introduce *coregionalised locomotion envelopes* - a method for multi-dimensional manifold regression, on human locomotion variates. Herein we render a qualitative description of this method.

We have previously considered novel control strategies for powered ankle-foot prostheses, using a data-driven approach, which employs a combination of Gaussian processes regression and impedance control [2]. Therein we learned the nonlinear functions, which dictate how locomotion variables temporally evolve using the aforementioned nonparametric method, and regress that surface over several velocities to create a manifold, per variable. The joint set of manifolds, as well as the temporal evolution of the gait-cycle duration is what we termed a locomotion envelope. For full notational and methodological details, please consult that paper, as we build upon it extensively herein and will not repeat background material.

The problem with our initial approach is that it did not consider the dependency (or correlation) of the input training variables or the output training variables, instead, assuming that regression variates are independent (strictly i.i.d.). Our initial approach is not an uncommon first pass. Indeed, consider the remarks by Mai & Commuri [4] on the topic of commercially available below-knee controlled prostheses:

> In general, control algorithms neglect the dynamics of the ankle joint, the interaction of the ankle with the remaining healthy joints of the residual limb, and the effect of the ground reaction torque. These devices are based on the linearized dynamics of the joint and use proportional-derivative control with fixed control gains.

---

[*]Corresponding author.

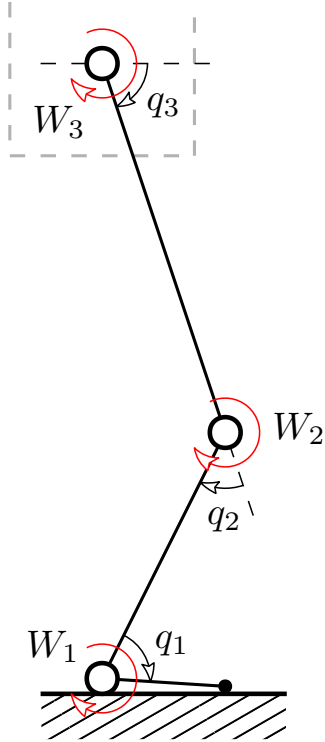

Figure 1: Rigid body diagram for lower-body functions, seen here in the sagittal plane. Active degrees of freedom are given as $\{q_i \mid i = 1, 2, 3\}$ and the dependent variable is the work given by $\{W_j \mid j = 1, 2, 3\}$.

Given the strongly dependent nature of human locomotion, in everything from the cyclical nature of exogenously measured metrics (such as the angular velocity of the knee-angle – see fig. 1), to the proprioceptive nature of our indodgenous control mechanisms, there is ample evidence to suggest that control variates *are* dependent. This strong assumption will form the basis for this future direction; a sequel to our initial contribution which takes into account human locomotion correlates.

The issue which we seek to tackle is similar to the *inverse dynamics problem* for robotic manipulators. In that domain, we seek to compute the torques $\tau$ required at the joints, to drive a manipulator along some given trajectory. That trajectory could e.g. be specified as the temporal evolution of the joint-angles; $\mathbf{q}(t)$, velocities; $\dot{\mathbf{q}}(t)$ and accelerations; $\ddot{\mathbf{q}}(t)$. See fig. 1 wherein $\mathbf{q} \triangleq [q_1, q_2, q_3]^\mathsf{T}$. In that familiar problem, it then becomes a quest to a find a model for $\tau(\ddot{\mathbf{q}}, \dot{\mathbf{q}}, \mathbf{q})$. As Williams et al. [6] note, this is hard enough in robotics proper, in our domain it is harder still. Not only do we have to contend with the uncertainty of the physical parameters of the robot (or in our case; the active prosthesis – not to mention the dynamical properties of the subject whose mass e.g. will change on a daily basis), we also need to model such things as human-robot interactions, ground friction, changing operational environments, human adaptability to the device and long-term rehabilitation – to mention but a few. Instead of torque, our dependent variable is the work $\mathbf{W} \triangleq [W_1, W_2, W_3]^\mathsf{T}$, resulting from the velocity of locomotion $v$. Where, to echo the study by Williams et al. [6], a *context* in our study refers to the different velocities we seek to regress over, such that the inverse dynamics function depends on the different contexts. To find this field of functions we can turn to Gaussian process regression, but this time for vector valued outputs as shown by the graphical model in fig. 2.

## 1 Vector valued outputs

In the classical supervised setting, the learning problem considers the estimation of the output, for any given input $\mathbf{x}_*$. The output is estimated through an estimator $y_* = f_*(\mathbf{x}_*)$, which takes $N$

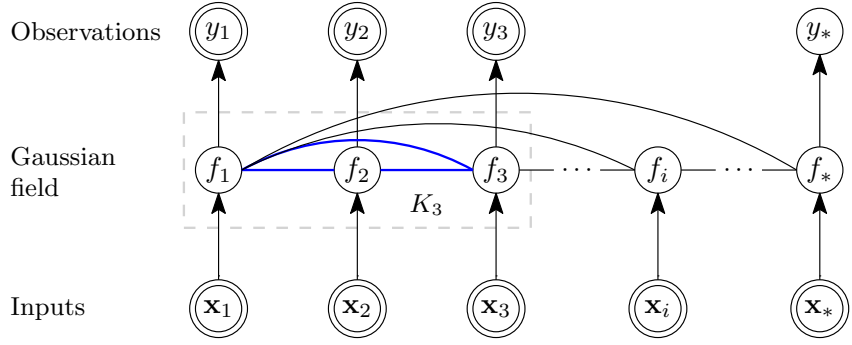

(a) Gaussian process graphical model.

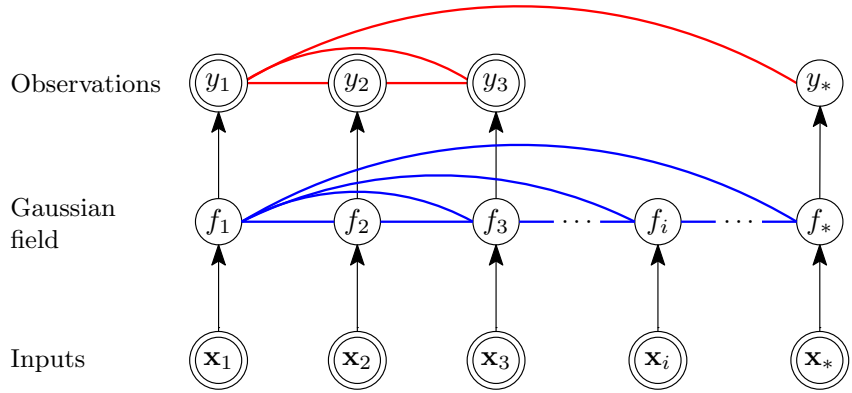

(b) Coregionalised Gaussian process graphical model.

Figure 2: Gaussian process graphical model. All hidden nodes $f_i = f(\mathbf{x}_i)$ (for $i = 1, 2, 3$) are interconnected by directed edges forming the Gaussian process. For brevity, and to avoid clutter, we only show conditional dependencies of $f_1$ s.t. $f_1 \mid f_2, f_3, f_i, f_*$. Adapted from [5, p. 17]. Special care should be taken regarding the stochastic nature of the observed nodes. Specifically note that the inputs are deterministic i.e. under this model, but the observations are stochastic per design through the Gaussian process, as they are sampled.

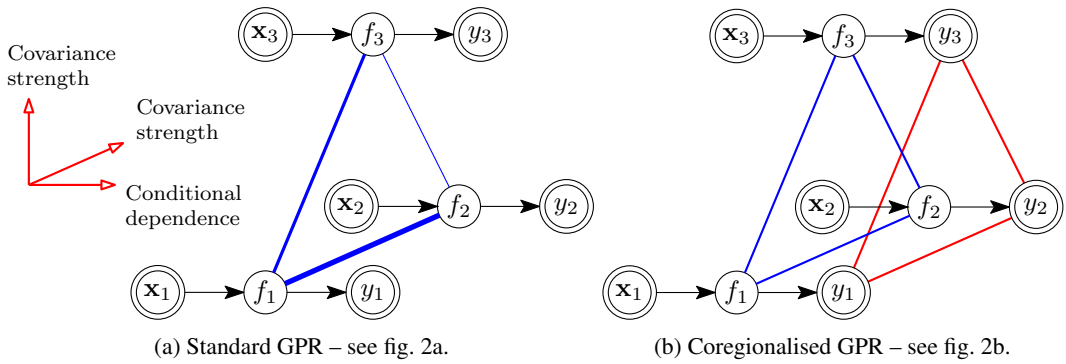

(a) Standard GPR – see fig. 2a.    (b) Coregionalised GPR – see fig. 2b.

Figure 3: Oblique projection of Gaussian process $K$-complete (nomenclature borrowed from graph theory) sub-graphical models with and without test points. Note that edges carry different weights, corresponding to different strengths of covariance. The 'vertical' and 'horizontal' strengths (as noted by coordinate system) are in reference to other nodes that lie in the same plane. The final, orthogonal axis, of the coordinate system demonstrates conditional dependencies or alternatively, information flow, in the sub-graphs. This depiction makes reference to fig. 2.

number of input-output pairs $\mathbf{S} \triangleq (\mathbf{X}, \mathbf{Y}) \equiv \{(\mathbf{x}_1, y_1), (\mathbf{x}_2, y_2), \ldots, (\mathbf{x}_N, y_N)\}$. Alvarez et al. [1] explain that the input space $\mathbf{x} \in \mathcal{X}$ is usually a space of vectors, whilst the output space $y \in \mathcal{Y}$ is typically a space of scalars (in our case, it is $v \in \mathbb{R}^+$). In multiple output learning, our focus $\mathcal{Y}$, is a space of vectors and the estimator $f_*(\cdot)$ is now a vector valued function $\mathbf{f}$, such that

$$\mathbf{f} \sim \mathcal{GP}(\mathbf{m}, \mathbf{K}) \tag{1}$$

in which $\mathbf{m}$ is a vector of mean functions and $\mathbf{K}$ is a positive matrix valued function. This can also be construed as the problem of solving $D$ distinct supervised problems, where each problem is modelled by one of the components $\{f_1, \ldots, f_D\} \in \mathbf{f}$. Another way, more relevant for our end purposes, is that of regression, in which case the likelihood function for the outputs is given by

$$\mathbb{P}(\mathbf{y} \mid \mathbf{f}, \mathbf{x}, \Sigma) = \mathcal{N}(\mathbf{f}(\mathbf{x}), \Sigma) \tag{2}$$

where $\Sigma \in \mathbb{R}^{D \times D}$ is a diagonal matrix. Then for a Gaussian likelihood, the predictive distribution and the marginal likelihood can be derived analytically [1], where the former is given as

$$\mathbb{P}(\mathbf{f}_* \mid \mathbf{S}, \mathbf{f}, \mathbf{x}_*, \phi) = \mathcal{N}(\mathbf{f}_*(\mathbf{x}), \mathbf{K}_*(\mathbf{x}_*, \mathbf{x}_*)) \tag{3}$$

in which

$$\mathbf{f}_*(\mathbf{x}_*) = \mathbf{K}_*^\mathsf{T}(\mathbf{K}(\mathbf{X}, \mathbf{X}) + \boldsymbol{\Sigma})^{-1}\overline{\mathbf{y}}$$
$$\mathbf{K}_*(\mathbf{x}_*, \mathbf{x}_*) = \mathbf{K}(\mathbf{x}_*, \mathbf{x}_*) - \mathbf{K}_{\mathbf{x}_*}(\mathbf{K}(\mathbf{X}, \mathbf{X}) + \boldsymbol{\Sigma})^{-1}\mathbf{K}_{\mathbf{x}_*}^\mathsf{T}$$

where $\boldsymbol{\Sigma} = \Sigma \otimes \mathbf{I}_N$ and $\mathbf{K}_{\mathbf{x}_*} \in \mathbb{R}^{D \times ND}$, such that $\phi$ denotes a possible set of hyperparameters of the covariance function $\mathbf{K}(\mathbf{x}, \mathbf{x}')$ used to compute $\mathbf{K}(\mathbf{X}, \mathbf{X})$ [1]. Note that $\otimes$ represents the Kronecker product between matrices.

The problem is now quite well defined as noted by Alvarez et al. [1]; the key insight is to work under the assumption that the problems are in some way related. Upon which the idea is to exploit the relation among the problems to improve upon solving each problem separately. To be more precise, we posit that the work required to turn each joint in $\mathbf{q}$, i.e. $W_h \; \forall h \in \mathcal{H}$, have members all of which are correlated. Specifically $W_h(t) \mid W_{-h}(t) \; \forall h \in \mathcal{H}$, where we use notation $W_{-h} \triangleq \{W_1, \ldots, W_{h-1}, W_{h+1}, \ldots, W_{|\mathcal{H}|}\}$. Thus we seek a vector field $\mathbf{f}$ which relates our inputs to our outputs, but which also takes into account the covariance between all the outputs $\mathbf{W}$. Such a model is specified by *coregionalised* GPs.

## 2 Coregionalised Gaussian processes

The coregionalised regression model relies upon the use of vector-valued kernels, one of the most common approaches for this regression is the linear model of coregionalisation (LMC) in which the outputs are expressed as linear combinations of independent random functions [1, §4.2]. In LMC the covariance function $\mathbf{K}(\mathbf{x}, \mathbf{x}')$ is given by

$$\mathbf{K}(\mathbf{x}, \mathbf{x}') = \sum_{q=1}^{Q} \mathbf{B}_q k_q(\mathbf{x}, \mathbf{x}') \tag{4}$$

here $\mathbf{B}_q \in \mathbb{R}^{D \times D}$ is the coregionalisation matrix for basis function $q$. This sum of separable kernels (i.e. input and output are decoupled) "represents the covariance function as the sum of the products of two covariance functions, one that models the dependence between the outputs" [1, §4.2.1], and the other which does the same thing but for the inputs. The coregionalised regression model is shown in fig. 3 where it is also compared to traditional GP regression. Note that in fig. 3 the Gaussian field is a complete graph $K_n$, where a complete graph is a simple undirected graph in which every pair of distinct vertices $n$ is connected by a unique edge. For our purposes, the edge strengths represent the covariance terms $\sum_{ij} = k(\mathbf{x}_i, \mathbf{x}_j)$.

A simpler version of LCM known as the "intrinsic coregionalisation model" (ICM) [1], was used by Williams et al. [6] for multi-task learning in the context of armed robots. But, noted Goovaerts [3], this model is more restrictive than the LMC because it assumes that each $k_q(\mathbf{x}, \mathbf{x}')$ contributes equally to the "construction of the autocovariances and cross covariances for the outputs" [1]. This drawback is easy to fathom when the essential difference between LCM and ICM is that for the latter $Q = 1$, whereas we can specify any number of latent functions for the LCM.

## 3 Coregionalised locomotion envelopes

As we have discussed, the LCM and ICM share information across the outputs. This is something independent models cannot do. In the regions where there are no training observations specific to an output, the independent models tend to return to the prior assumptions, as is inherent in the model. But in the case where outputs have associated patterns, the fit is better with the coregionalised models, because they share information regarding the evolution of the outputs. This is a pertinent observation for coregionalised locomotion envelopes, since this is precisely the scenario where independent models may not be the most appropriate. Because of the cyclical nature of human motion, we can incorporate our very strong assumptions into the LCM, and regress over variates where we cannot easily (or e.g. due to the very high cost of doing so) collect observations.

Furthermore, because we eventually seek to perform transfer learning between subjects (i.e. using training observations from healthy subjects and apply them to a prosthesis for a disabled subject, which could potentially allow the latter to operate locomotion in a somewhat near natural manner) the LCM approach is still more suitable than independent models. Because not only is information shared between all the members in $\mathcal{Y}$, we can enforce that information is also shared between output spaces $\mathcal{Y}_s$ themselves. Where we index a subject-specific output space by $s$. This will be beneficial when input-output pairs $\mathbf{S}$ do not precisely match between subjects, or if there are training segments or contexts, missing for one subject, but which exist for another.